\documentclass[fleqn,twocolumn,10pt]{article}
\usepackage{cmff}
\usepackage{siunitx}
\usepackage{tikz}
\usetikzlibrary{calc}

\title{Predicting the flow field in a U-bend with deep neural networks}
\author{
    \textsf{Gergely HAJGAT\'{O}}\affiliation{1},
        \textsf{B\'{a}lint GYIRES-T\'{O}TH}\affiliation{2},
        \textsf{Gy\"{o}rgy PA\'{A}L}\affiliation{3}
}
\address{1}{Corresponding Author. Department of Hydrodynamic Systems, Faculty of Mechanical Engineering, Budapest University of Technology and Economics. M\H{u}egyetem rkp. 3, H-1111 Budapest, Hungary. Tel.: +36 1 463 3097, E-mail: ghajgato@hds.bme.hu}
\address{2}{Department of Telecommunications and Media Informatics, Faculty of Electrical Engineering and Informatics, Budapest University of Technology and Economics. E-mail: toth.b@tmit.bme.hu}
\address{3}{Department of Hydrodynamic Systems, Faculty of Mechanical Engineering, Budapest University of Technology and Economics. E-mail: gypaal@hds.bme.hu}

\begin{document}
\maketitle
\begin{abstract}
This paper describes a study based on computational fluid dynamics (CFD) and deep neural networks that focusing on predicting the flow field in differently distorted U-shaped pipes. The main motivation of this work was to get an insight about the justification of the deep learning paradigm in hydrodynamic hull optimisation processes that heavily depend on computing turbulent flow fields and that could be accelerated with models like the one presented. The speed-up can be even several orders of magnitude by surrogating the CFD model with a deep convolutional neural network.

An automated geometry creation and evaluation process was set up to generate differently shaped two-dimensional U-bends and to carry out CFD simulation on them. This process resulted in a database with different geometries and the corresponding flow fields (2-dimensional velocity distribution), both represented on 128x128 equidistant grids. This database was used to train an encoder-decoder style deep convolutional neural network to predict the velocity distribution from the geometry.

The effect of two different representations of the geometry (binary image and signed distance function) on the predictions was examined, both models gave acceptable predictions with a speed-up of two orders of magnitude.
\end{abstract}

\keywords{convolutional neural networks, deep learning, deep neural networks, flow field predicition, metamodelling, surrogate model}

\printnomenclature

\section{Introduction}
The role of machine learning (and particularly deep learning) evolved in the recent years in many research fields that were mastered by highly qualified human experts traditionally. The actual trends and future possibilities are summarised by the authors of \cite{theoryguided}. As computational resources keep growing, even larger tasks can be aided or completely surrogated by deep neural networks (DNN). Computational fluid dynamics (CFD) simulations are typical time-consuming processes in the area of fluid-dynamic design. The simulations are necessary to analyse a new idea but are resource-intensive as well, and the common numerical methods in commercial software are weakly parallelisable compared to DNNs. These drawbacks are amplified when hundreds of CFD simulations have to be computed, e.g. during an optimisation session. The need for appropriate surrogate models is growing, as optimisation-driven design is spreading.

Satisfying results were achieved by Beigzadeh et al. \cite{rahimi}, Verstraete et al. \cite{vki} and Duvigneau and Visonneau \cite{duvigneau} among others in surrogating CFD simulations during optimisation. Each of these solutions tries to predict the value of some integral quantity (e.g. Nusselt number, pressure loss over the domain, etc.) from some set of parameters that are modified step by step by an evolutional algorithm. Although this approach gives satisfying results in most cases, they depend on the parametrisation and microscopic information in the flow domain vanishes.

Guo et al. \cite{xiao} present the state-of-the-art result in this topic by predicting the complete velocity field from a parameter-free description of the geometry. Although the topology of the neural network is more complicated compared to \cite{rahimi}, \cite{vki} and \cite{duvigneau}, it nearly provides the same amount of information that would be delivered by the accurate CFD simulation. The convolutional neural network (CNN) used in \cite{xiao} was trained with laminar, external flows; its ability to predict the flow field under turbulent flow conditions has not been tested yet. Although the results were satisfactory, there is only a limited application area where fluid flows can be viewed as laminar. 

Besides, several studies are present in the literature in the topic of machine learning and turbulence. Singh et al. \cite{wingman} try to improve a simple turbulence model to get better CFD simulation results in the flow fields computed around airfoils. In this case, the knowledge-based model is augmented with a part that is acquired by machine learning techniques. In contrast, the authors of \cite{discrep} keep the former knowledge-based model intact and machine learning is used to predict the discrepancy of the model from accurate simulation data. In \cite{invari} deep learning techniques are utilized instead of conventional machine learning techniques. The authors aim at predicting the anisotropic properties of turbulence with granting Galilean invariance at the same time. Lguensat et al. \cite{ocean} detected large-scale eddies by DNNs. These studies show that machine learning techniques have the potential to predict certain properties of turbulence.

This paper presents a convolutional neural network inspired by Guo et al. \cite{xiao}, wherewith the velocity field can be predicted for internal, turbulent flows under specific boundary conditions and in a given geometry with a bearable error. The speed-up is two orders of magnitude compared to the CFD simulation if only one geometry has to be evaluated. The benefit would grow further by increasing the number of geometries to evaluate, e.g. in optimisation sessions.

\section{Neural networks as surrogate models}
Feed-forward artificial neural networks are universal function approximators as stated by Hornik et al. \cite{hornik} and Cybenko \cite{cybenko}. Although the model is biologically inspired, it is superfluous to examine the neuron nerve cell system of an animal to get an understanding of artificial neural networks. Moreover, artificial and real neural networks are similar only from a sufficiently large distance of view.

The topology of a neural network is flexible enough to get it tailored to specific problems, thus many types of building blocks exist. The present work relies on multi-layer perceptrons and convolutional layers whose basics are described here.

\subsection{Multi-layer perceptron}
One of the basic applications of neural networks is the prediction of valuable information from the corresponding features of an entity. To accomplish this goal, the features are quantified and multiplied with a specific weight that measures the importance of that feature. The sum of these products is the signal that is fed to an activation function, which can be, e.g. sigmoid, hyperbolic tangent or rectified linear unit \cite{relu}. Considering that the weights are known, the output of the activation function is suitable for classification or regression.

\begin{figure}[h]
    \centering
    \includegraphics[width=.65\linewidth, keepaspectratio]{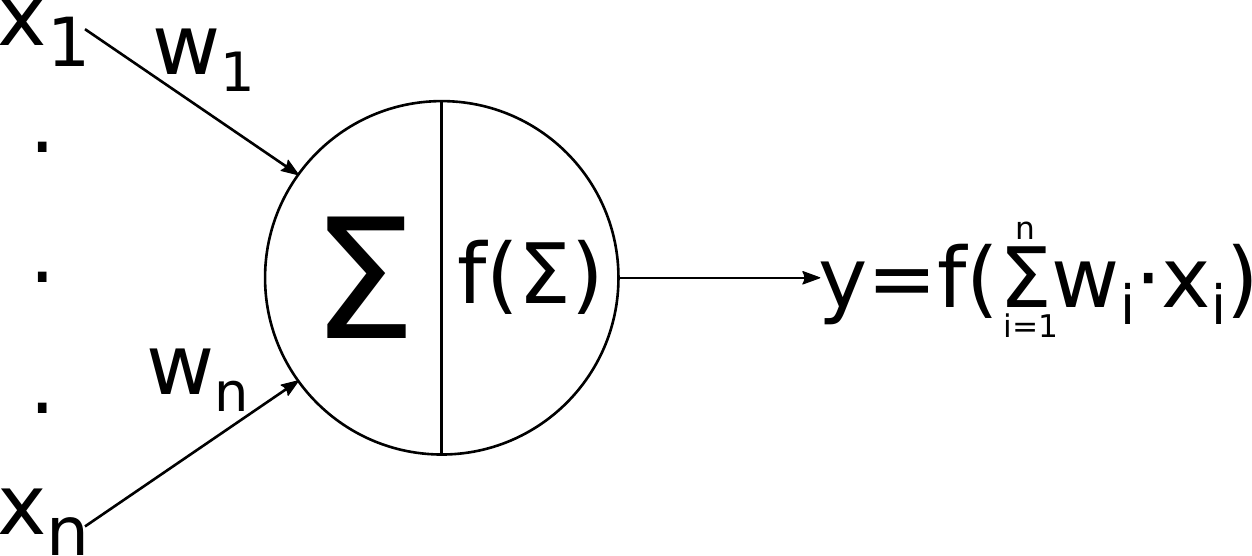}
    \caption{A single perceptron, where $x_i$ is the $i$th feature of the entity and $w_i$ is the weight belonging to the $i$th feature.}
    \label{perceptron}
\end{figure}

This simple setup is called perceptron\footnote{The original perceptron uses sigmoid activation function.} (depicted in Figure~\ref{perceptron}), while the combination of the summation and the activation function is called neuron. As Minsky \cite{minsky} stated, a standalone perceptron is unable to cope with situations when the features have to model exclusive-or logic. To overcome this drawback, neurons are arranged into layers and stacking these layers results in the multi-layer perceptron. 

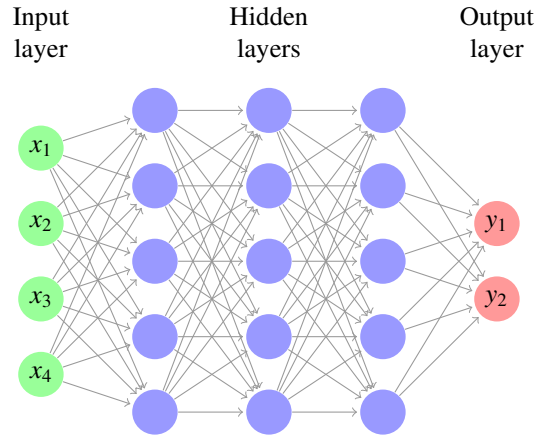
\begin{figure}
    \centering
    \def\layersep{1.5cm}
    \begin{tikzpicture}[shorten >=1pt,->,draw=black!40, node distance=\layersep]
        \tikzstyle{every pin edge}=[<-,shorten <=1pt]
        \tikzstyle{neuron}=[circle,fill=black!25,minimum size=17pt,inner sep=0pt]
        \tikzstyle{input neuron}=[neuron, fill=green!40];
        \tikzstyle{output neuron}=[neuron, fill=red!40];
        \tikzstyle{hidden neuron}=[neuron, fill=blue!40];
        \tikzstyle{annot} = [text width=4em, text centered]
    
        \foreach \name / \y in {1,...,4}
            \node[input neuron] (I-\name) at (0,-\y) {$x_\y$};
    
        \foreach \name / \y in {1,...,5}
            \path[yshift=0.5cm]
                node[hidden neuron] (Hi-\name) at (\layersep,-\y cm) {};
        \foreach \name / \y in {1,...,5}
            \path[yshift=0.5cm]
                node[hidden neuron] (Hii-\name) at (2*\layersep,-\y cm) {};
        \foreach \name / \y in {1,...,5}
            \path[yshift=0.5cm]
                node[hidden neuron] (Hiii-\name) at (3*\layersep,-\y cm) {};
    
        \foreach \name / \y in {1,...,2}
            \path[yshift=-1.0cm]
                node[output neuron] (O-\name) at (4*\layersep,-\y cm) {$y_\y$};
    
        \foreach \source in {1,...,4}
            \foreach \dest in {1,...,5}
                \path (I-\source) edge (Hi-\dest);
        \foreach \source in {1,...,5}
            \foreach \dest in {1,...,5}
                \path (Hi-\source) edge (Hii-\dest);
        \foreach \source in {1,...,5}
            \foreach \dest in {1,...,5}
                \path (Hii-\source) edge (Hiii-\dest);
        \foreach \source in {1,...,5}
            \foreach \dest in {1,...,2}
                \path (Hiii-\source) edge (O-\dest);
        \node[annot,above of=I-1, node distance=1.5cm] (hl) {Input layer};
        \node[annot,above of=Hii-1, node distance=1cm] (hl) {Hidden layers};
        \node[annot,above of=O-1, node distance=2.5cm] (hl) {Output layer};
    \end{tikzpicture}

    \caption{Example of a multi-layer perceptron with 4 input features, 3 hidden layers and 2 outputs.}
    \label{dense}
\end{figure}

The neurons in a topology described above can be organized as shown in Figure~\ref{dense}. The first layer is called the input layer, thereafter come the so-called hidden layers and finally the output layer. As all of the neurons on a layer are connected to all of the neurons of the next layer, this topology is also called densely or fully connected.
When these layers are stacked different level of abstractions are presented in different depths of a proper network topology that is appropriately trained \cite{nature}. This is the reason why it is a common practice to apply many layers for complex tasks that lead to deep neural networks and hence comes the name deep learning.

To be able to give good predictions, the weights of the neurons have to be properly aligned. The appropriate weights are computed during the training of the network that is an iterative process and is not discussed here. The interested readers are encouraged to read the related chapters of the book of Yaser et al. \cite{yaser} and the work of LeCun et al. in \cite{lecun}.

\subsection{Convolutional layers}
\begin{figure}[h]
    \centering
    \def\sep{.5}
    \def\filterybias{-.5}
    \begin{tikzpicture}[shorten >=1pt,->,draw=blue!50, node distance=\sep]
        \tikzstyle{every pin edge}=[<-,shorten <=1pt]
        \tikzstyle{neuron}=[circle,draw=black!50,minimum size=13pt,inner sep=0pt]
        \tikzstyle{neuron emph}=[circle,draw=black,minimum size=13pt,inner sep=0pt]
        \tikzstyle{neuron filter}=[circle,draw=blue,minimum size=13pt,inner sep=0pt]
        \tikzstyle{annot} = [text width=4em, text centered]
        \tikzset{between/.style args={#1 and #2}{at = ($(#1)!0.5!(#2)$)}}
    
        \foreach \name / \y in {1,...,2}
            \node[neuron emph] (d1-\name) at (0,-.5*\y) {};
        \foreach \name / \y in {3,...,4}
            \node[neuron] (d1-\name) at (0,-.5*\y) {};
        \foreach \name / \y in {1,...,2}
            \node[neuron emph] (d2-\name) at (\sep,-.5*\y) {};
        \foreach \name / \y in {3,...,4}
            \node[neuron] (d2-\name) at (\sep,-.5*\y) {};
        \foreach \name / \y in {1,...,4}
            \node[neuron] (d3-\name) at (2*\sep,-.5*\y) {};
        \foreach \name / \y in {1,...,4}
            \node[neuron] (d4-\name) at (3*\sep,-.5*\y) {};
    
        \foreach \name / \y in {1,...,2}
            \node[neuron filter] (f1-\name) at (5*\sep,\filterybias-.5*\y) {};
        \foreach \name / \y in {1,...,2}
            \node[neuron filter] (f2-\name) at (6*\sep,\filterybias-.5*\y) {};
    
        \node[neuron emph] (o1-1) at (8*\sep,-.5) {};
        \foreach \name / \y in {2,...,3}
            \node[neuron] (o1-\name) at (8*\sep,-.5*\y) {};
        \foreach \name / \y in {1,...,3}
            \node[neuron] (o2-\name) at (9*\sep,-.5*\y) {};
        \foreach \name / \y in {1,...,3}
            \node[neuron] (o3-\name) at (10*\sep,-.5*\y) {};
    
        \path (d1-1) edge (f1-1);
        \path (d1-2) edge (f1-2);
        \path (d2-1) edge (f2-1);
        \path (d2-2) edge (f2-2);
        
        \path (f1-1) edge (o1-1);
        \path (f1-2) edge (o1-1);
        \path (f2-1) edge (o1-1);
        \path (f2-2) edge (o1-1);
    
        \node[annot,between=d2-1 and d3-1](auxi){};
        \node[annot,between=f1-1 and f2-1](auxf){};
    
        \node[annot,above of=auxi, node distance=.5cm] (inp) {Input};
        \node[annot,above of=auxf, node distance=1cm] (fil) {Filter};
        \node[annot,above of=o2-1, node distance=.5cm] (out) {Output};
    \end{tikzpicture}
    \caption{Simplified example of a convolutional layer.}
    \label{cnn}
\end{figure}
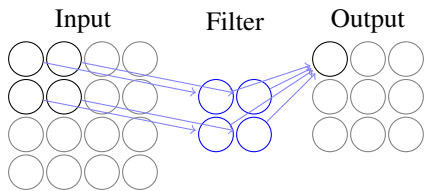

Densely connected layers are inefficient in pattern recognition tasks because they do not have a sense on the data spatiality. Convolutional layers learn (during training) and apply (during inference) filters on the input (or on the preceding layer).
These filters are cross-correlated with the input, so the output of a filter is the sum of the Hadamard product of the filter and the appropriate parcel of the data where the filter is applied. This leads to the reduction of the dimensionality as depicted in Figure~\ref{cnn}. As the size of the filter is intended to be much smaller as the size of the input layer the filter has to be slid over the input. The stride of a filter counts the number of elements (pixels, neurons, etc.) wherewith the filter is slid over the domain during cross-correlation. In such cases when the filter size with the prescribed stride does not match the size of the data, the original data is augmented and the size of the augmentation is called padding.

Training and applying filters on the preceding layer is conventionally called convolution in deep learning and is very effective in representation learning, and thus, pattern recognition tasks even when images or sequential data are the origin \cite{lenet}. Deep convolutional neural networks are the state-of-the-art solutions to object detection. See e.g. the ImageNet Challenge winner works of Krizhevsky et al. \cite{alexnet12}, Szegedy et al. \cite{googlenet14}. and He et al. \cite{residu}.

\subsection{Motivation}
The present work was motivated by the results achieved recently by deep convolutional neural networks in pattern recognition and object detection. Guo et al. \cite{xiao} used a deep convolutional neural network to predict the laminar flow field ,,seeing'' only the geometry of the fluid domain. The present work relies on the assumption that convolutional networks have the power to recognize turbulent structures in the flow field and thus make acceptable predictions of the velocity field in turbulent internal flows as well.

\section{Methodology}
There are no preprocessed datasets publicly available wherewith convolutional neural networks could be trained to predict velocity fields in turbulent flows. Therefore, the data generation and the preprocessing were also part of the present work. This chapter describes the numerical model of the basic fluid dynamics problem. The preprocessing of the data wherewith the convolutional neural network was fed and the topology of the CNN itself was the result of a hyperparameter optimisation. The utilized software and packages are also mentioned in the last subsection.

\subsection{Fluid dynamics problem}\label{cfd}
The fluid dynamics problem is based on the one presented by Verstraete et al. in \cite{vki}. The original paper focuses on the shape optimisation of the geometry, thus the authors of \cite{vki} connected a CFD model with an evolutional algorithm to find the shape with the least pressure loss. The authors of the present article rebuilt the CFD model according to \cite{vki} but it was used to generate random shapes wherewith the convolutional neural network could be trained in this case. Besides, the different objective permitted minor simplifications in the simulation.

\subsubsection{Geometry}
\begin{figure}[h]
    \centering
    \includegraphics[height=.45\linewidth, keepaspectratio]{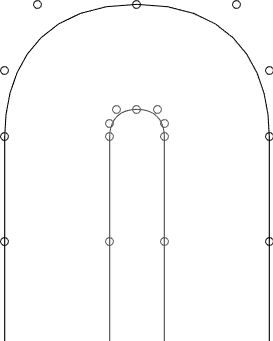}
    \hspace{0.5cm}
    \includegraphics[height=.45\linewidth, keepaspectratio]{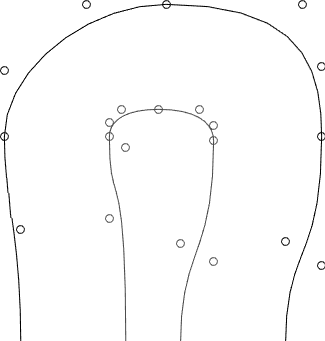}
    \caption{Planar view of the of the U-bend. Conventional shape in the left, a distorted shape in the right with circles denoting the endpoint tangents of Bezi\'{e}r curves.}
    \label{convent}
\end{figure}

The geometry is a curved channel that is a common part of cooling systems. The planar view of the conventional shape is depicted in Figure~\ref{convent}. This shape can be altered in the plane as the geometry is built up from B\'{e}zier-curves.\footnote{The actual parameterization can be found in \cite{vki}.} A distorted bend is also depicted in Fig.~\ref{convent}.

The spatial geometry would be the straight extrusion of the planar view in the \emph{z} dimension but the 3 dimensionality was omitted and the flow was considered planar. Additional straight parts on the up- and downstream sides were applied to handle the boundary conditions of the simulation in a proper way. The lengths of these legs were chosen such that the boundaries could not affect the flow pattern in the U-turn.

\subsubsection{Boundary conditions and simulation setup}
\begin{figure}[h]
    \centering
    \includegraphics[height=.85\linewidth, keepaspectratio]{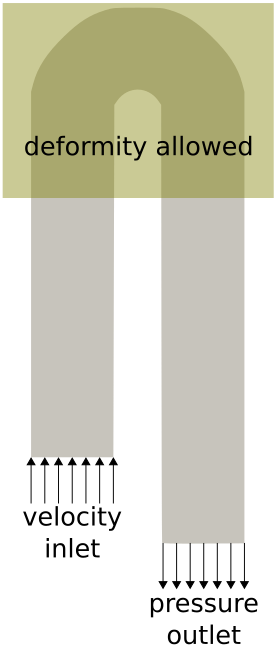}
    \caption{Position of inlet and outlet boundaries and the area where the geometry was a subject to change.}
    \label{boundary}
\end{figure}

In the physical model constant temperature, incompressible air with a kinematic viscosity of $1.5\cdot10^{-5}$ $m^2/s$ flowed through the channel. The medium entered the domain on the left side with a constant uniform velocity of $8.8$ $m/s$ and left it on the right side through  a constant pressure outlet of $0$ $Pa$ as shown in Fig.~\ref{boundary}. Although the velocity profile was uniform at the inlet, it got fully developed in the straight section before the curve.

The initial channel width was $75$ $mm$, but it was subject to change in the vicinity of the turn due to the parametrisation. The Reynolds number was $44000$ in the initial geometry which clearly classifies the flow as turbulent. Moreover, the fluid domain was bounded with hydraulically smooth walls, the bottom and top surfaces of the geometry were treated as symmetries.

The geometry was discretised with a structured numerical grid with near-wall refinement. The total number of elements is $14406$ and the resolution of the boundary layer was aligned to apply the use of the high-Re k-$\varepsilon$ turbulence model with the corresponding \emph{epsilonWallFunction} wall function of \verb+OpenFoam+. It also means that the anisotropy properties of turbulence were not considered in this study.

The numerical model was solved with the \emph{simpleFoam} solver of \verb+OpenFOAM+ with a convergence criterion of $10^{-4}$ for the residuals.

\subsection{Data processing}
The CFD simulation discussed in Section~\ref{cfd} was used to generate data for training the convolutional neural network. The CNN aimed at predicting the velocity field in a given geometry, thus these data had to be extracted from the CFD simulations. First, the data from the numerical grid of the CFD were converted to a rectangular, equidistant 128x128 mesh to make them applicable to the convolutional neural network. The subsequent steps were taken as follows.

\subsubsection{Geometry}
\cite{xiao} suggests two different kinds of geometry representation for predicting velocity fields around immersed bodies. Both of these methods were applied in the present work for internal flows.

\begin{figure}[h]
    \centering
    \includegraphics[width=.45\linewidth, keepaspectratio]{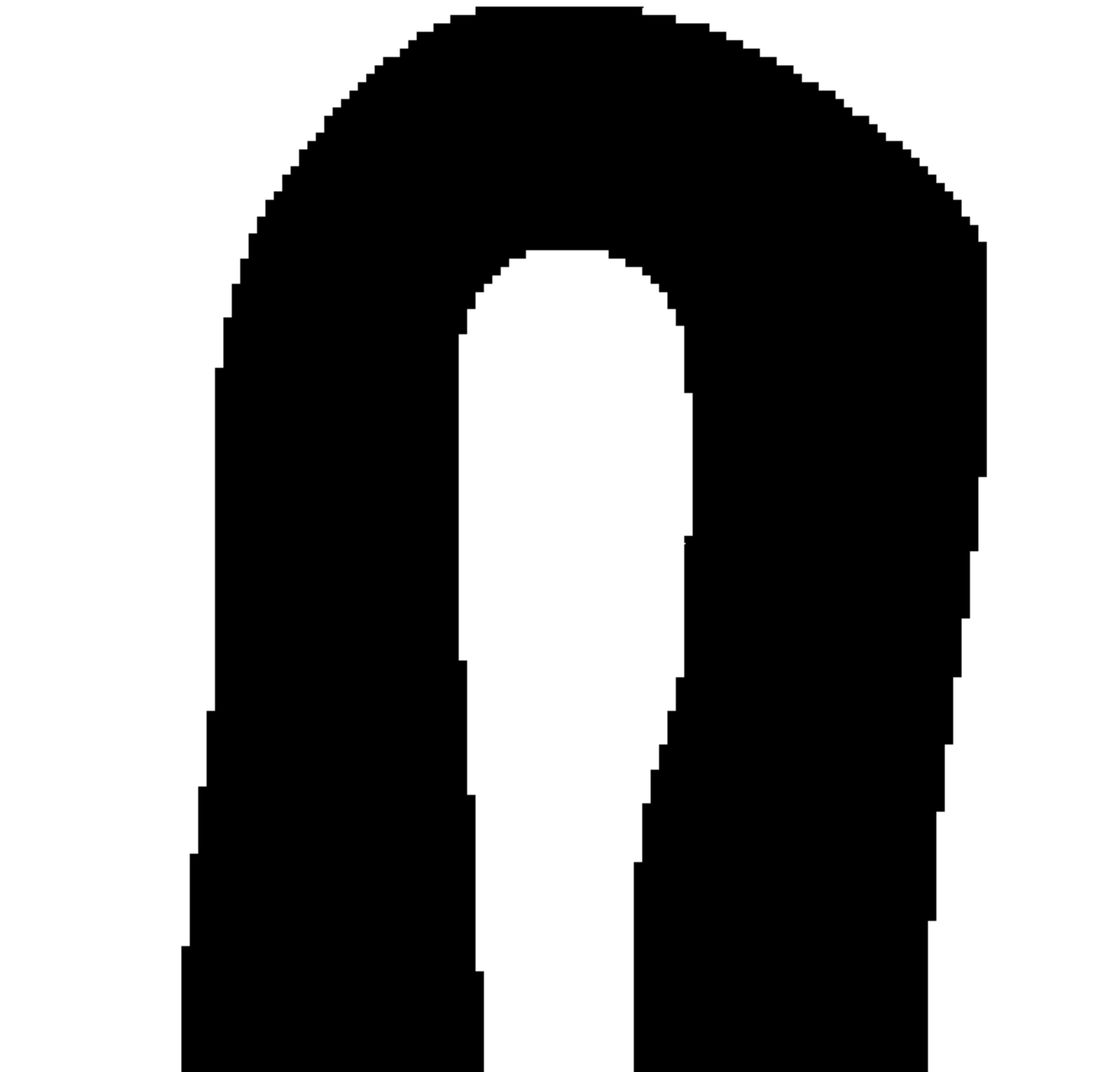}
    \includegraphics[width=.45\linewidth, keepaspectratio]{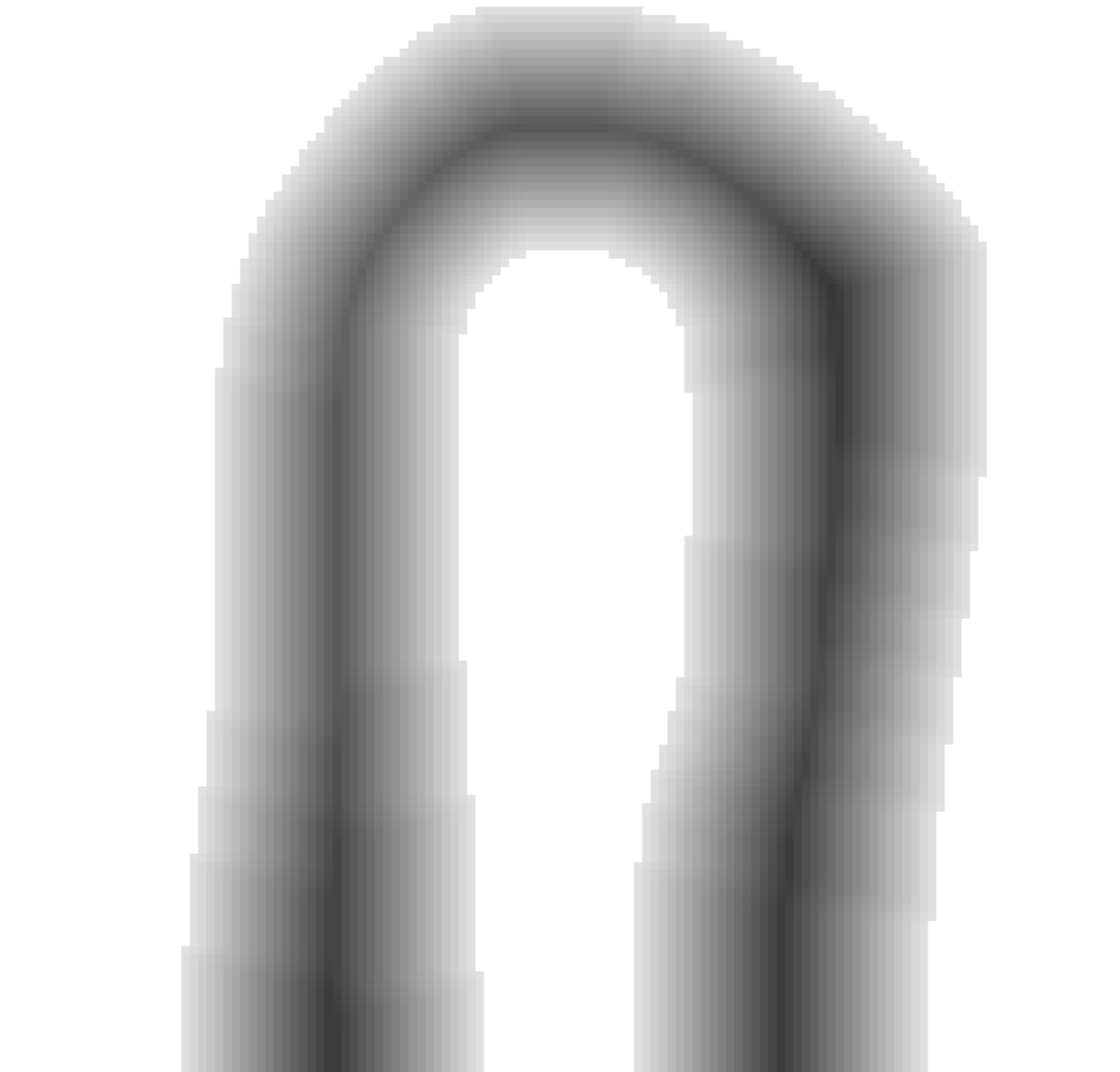}
    \caption{Geometry represented as binary image (left) and signed distance field (right).}
    \label{inputs}
\end{figure}

The simpler method is to convert the geometries to binary images, where the value of a grid node can be $1$ or $0$  only according to whether it is inside the fluid domain or not, respectively. A binary image is shown in the left side of Figure~\ref{inputs}.

A more sophisticated approach is to calculate the signed distance field in the fluid domain, that is a common technique in computer graphics, see e.g. \cite{compgraph}. \emph{Distance field} relates to a variable field where the value in a grid point refers to the distance between the actual point and the closest point on the boundary, while the term \emph{signed} means, that the sign of the value refers to whether the actual point is inside or outside the boundary. This approach was used with a slight modification here, as the signed distance function was positive inside the fluid domain and $0$ everywhere else.

The applied method is formulated in Eq.~\ref{sdf}, considering the fluid domain as a set $\Omega$ in the two-dimensional Euclidean space, $x$ and $y$ as arbitrarily chosen coordinates in the space and $x'$ and $y'$ as running variables on the boundary of the fluid domain.

A geometry converted to a signed distance field is depicted on the right hand side of Fig.~\ref{inputs} as well.
\begin{equation}
    f(x,y) =
        \begin{cases}
            \min\limits_{(x',y')\in\partial\Omega} |(x,y)-(x',y')|  & \mathrm{if}\ x,y \in\Omega\\
            0                                                       & \mathrm{if}\ x,y \in\overline{\Omega}
        \end{cases}
\label{sdf}
\end{equation}

\nomencNormal[50]{$\Omega$}{set in the two-dimensional Euclidean space}{$-$}

\nomencNormal[50]{$\overline\Omega$}{complement of set $\Omega$}{$-$}

\nomencNormal[50]{$\partial\Omega$}{boundary of $\Omega$}{$-$}

\nomencNormal[20]{$x$}{coordinate in the x direction}{$-$}

\nomencNormal[20]{$y$}{coordinate in the y direction}{$-$}

\nomencNormal[40]{$x'$}{running coordinate in the x direction}{$-$}

\nomencNormal[40]{$y'$}{running coordinate in the y direction}{$-$}

Both representations were available on a 128x128 equidistant mesh as mentioned earlier and images were standardised based on the data in the training set.

\subsubsection{Velocity field}
\begin{figure}[h]
    \centering
    \includegraphics[width=.45\linewidth, keepaspectratio]{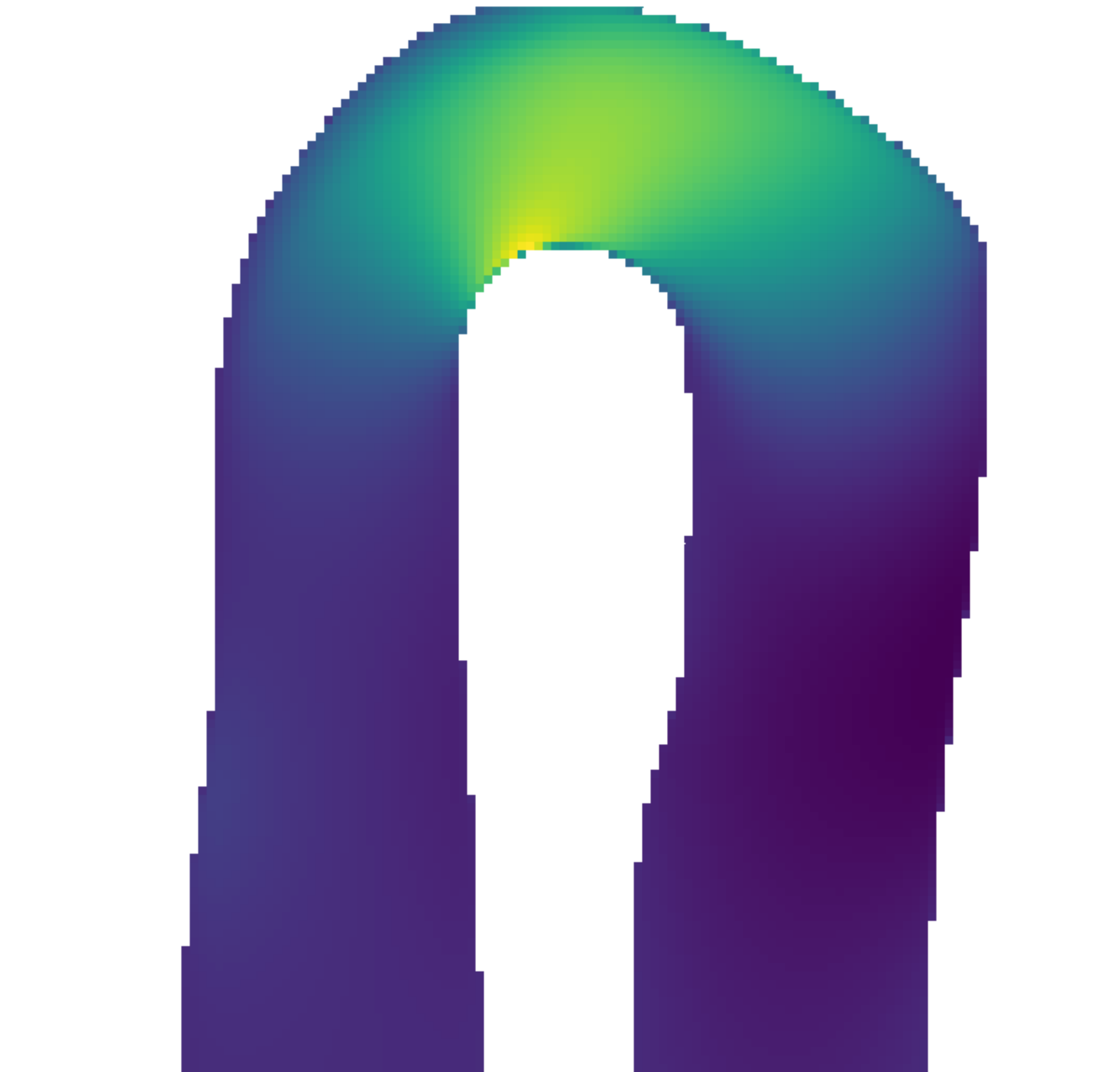}
    \includegraphics[width=.45\linewidth, keepaspectratio]{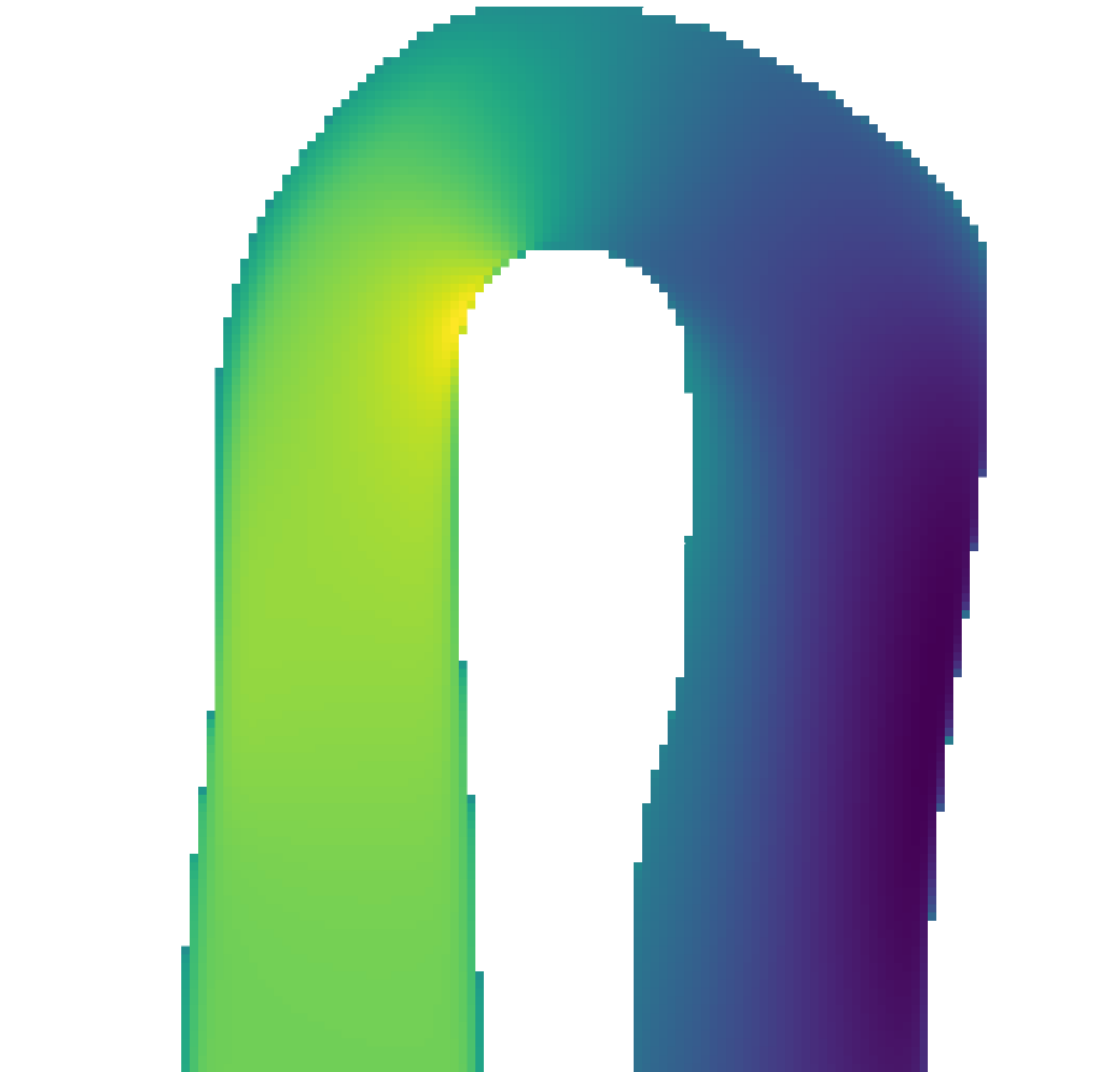}
    \caption{Velocity \emph{x} and \emph{y} components in the left and in the right, respectively. The figure serves as illustration only, thus scales were omitted.}
    \label{xyvelo}
\end{figure}

To make the velocity vectors reconstructable, the velocity fields in the \emph{x} and \emph{y} directions were extracted separately. Fields were resampled on the same equidistant, rectangular grid as the representations of the geometry with a resolution of 128x128 grid points. Sample velocity fields are depicted in Figure~\ref{xyvelo}. The background (region outside the fluid domain) was set to zero, but the scale is different for the $v_x$ and $v_y$ fields.

\subsection{Convolutional neural network}
\begin{figure*}[h]
    \centering
    \includegraphics[width=.95\textwidth, keepaspectratio]{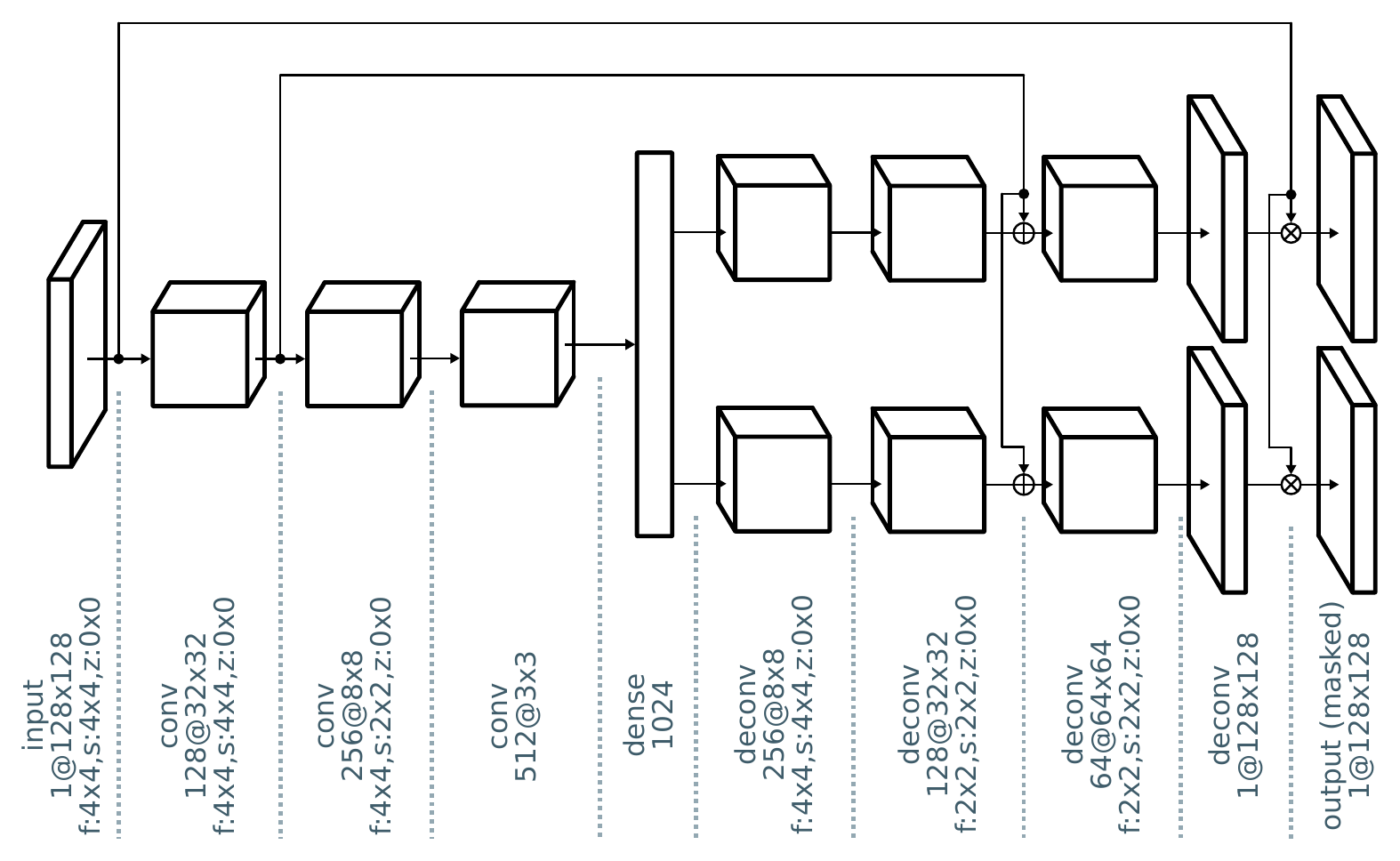}
    \caption{Architecture of the artifical neural network.}
    \label{convnet}
\end{figure*}

The convolutional neural network presented by Guo et al. \cite{xiao} was considered as a baseline for the present work but some parts of the network topology were subject to hyperparameter optimisation. The final architecture is depicted in Figure~\ref{convnet}. where \emph{conv} stands for convolution, \emph{deconv} for transposed convolution and \emph{dense} refers to a densely connected layer. The properties of the convolutional/deconvolutional layers are summarized as follows: \{depth\}@\{input\_x\}x\{input\_y\}, f:\{filter\_x\}x\{filter\_y\}, s:\{stride\_x\}x\{stride\_y\}, z:\{padding\_x\}x\{padding\_y\}. Please note that the filter properties of a convolutional/deconvolutional layer are denoted at the preceding layer since the filter is applied there. In case of the dense layer only the number of neurons are indicated.

This is an autoencoder-style neural network with convolutional-deconvolutional layers with a dense layer as a bottleneck. The encoder part takes the geometry represented as binary image or signed distance field on a 128x128 grid, while the outputs of the network are the standalone velocity fields in $x$ and $y$ directions. Thus, the decoder part is divided after the dense layer for predicting $v_x$ and $v_y$ fields. The output is masked with the binary image of the input to omit the prediction coming from the CNN outside of the fluid domain. It has been done so regardless of whether the input itself was a binary image or a signed distance field.

The error of the prediction was defined as the root mean squared error between the predicted and the CFD simulated (considered here as ground-truth) velocity fields. The mask was applied to the output of the CNN in the training phase as well, thus the network could not reach small prediction error values by predicting the constant field outside the fluid domain, but it was forced to make better predictions inside the fluid domain.

A residual connection is presented between the first convolutional and the second deconvolutional layer means, that the output of the first convolutional layer is added to the output of the second deconvolutional layer in each branch. This type of connection is discussed by He et al. in \cite{residu} as a technique to improve accuracy and convergence of CNNs.

The optimisation of the network's weights was carried out with Adam optimiser, introduced by Kingma and Ba in \cite{adam}.

\subsection{Technical details}\label{techdat}
The geometry was built and meshed in the freely accessible \verb+gmsh+ \cite{gmsh} mesher software, the flow pattern was solved with the open-source numerical solver \verb+OpenFOAM+, the postprocessing was carried out in \verb+ParaVIEW+ scripted in \verb+Python+.

Training and inference with neural network were done in \verb+Keras+ \cite{keras} deep learning framework, while hyperparameters were optimised with \verb+Hyperopt+ using the \verb+Hyperas+ frontend. Both the hyperparameter optimisation and the training were carried out on an NVIDIA Titan Xp GPU with on-line data augmentation running on an Intel i5-7500 CPU.

\section{Experiment setup and results}
\begin{figure*}[h]
    \centering
    \begin{tabular}{cccl}
        \includegraphics[width=.2\textwidth, keepaspectratio]{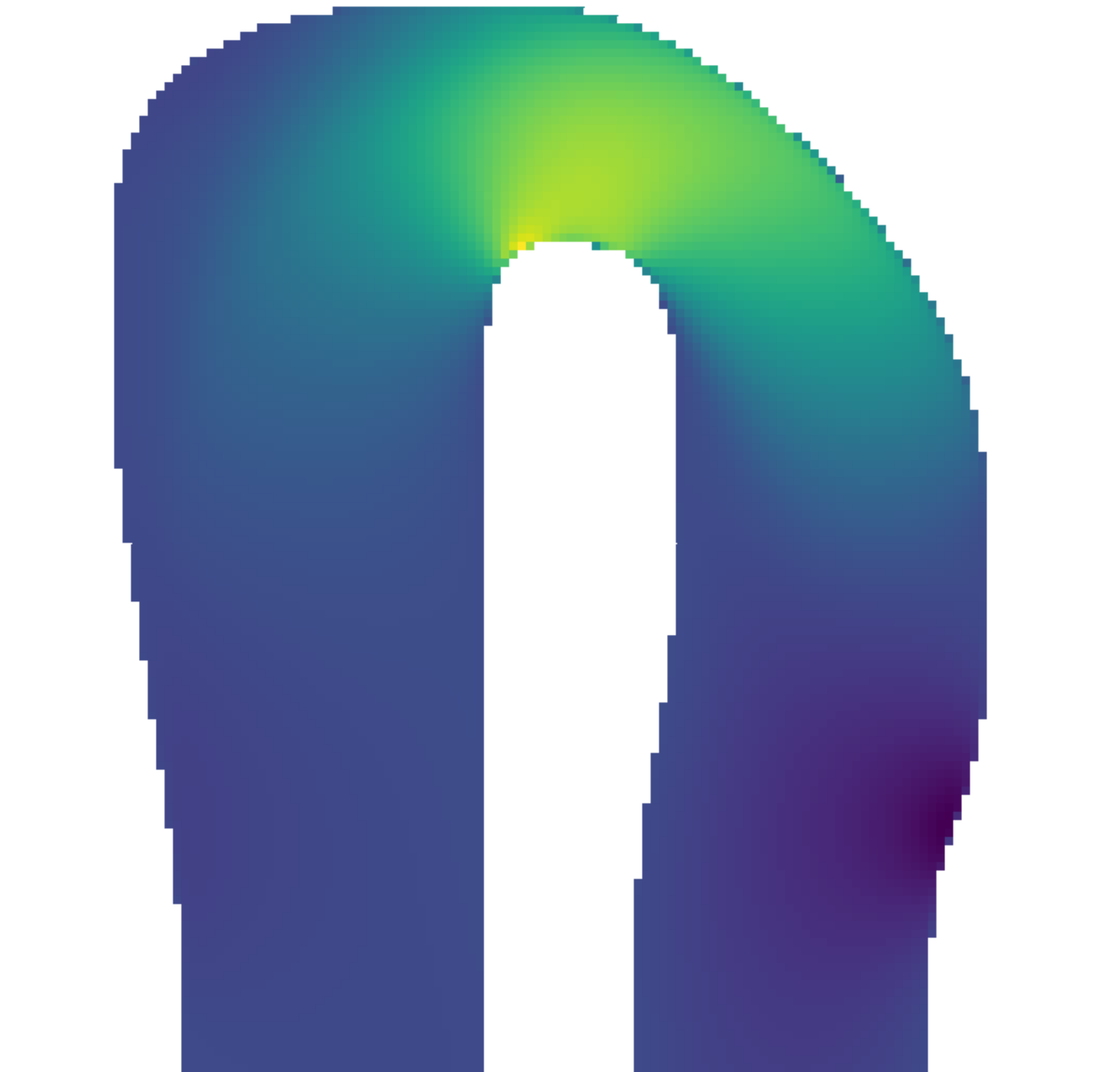}&
        \includegraphics[width=.2\textwidth, keepaspectratio]{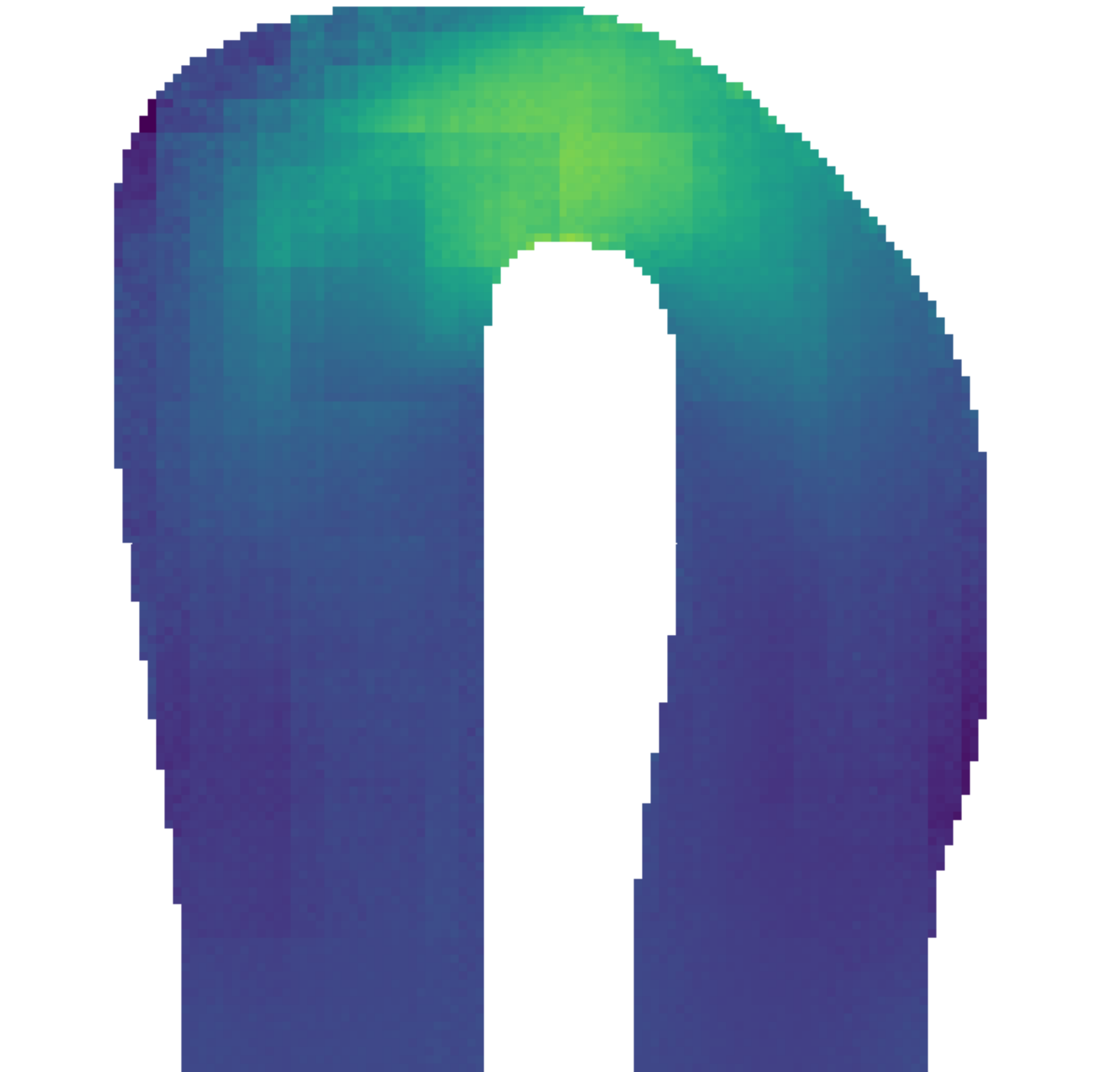}&
        \includegraphics[width=.2\textwidth, keepaspectratio]{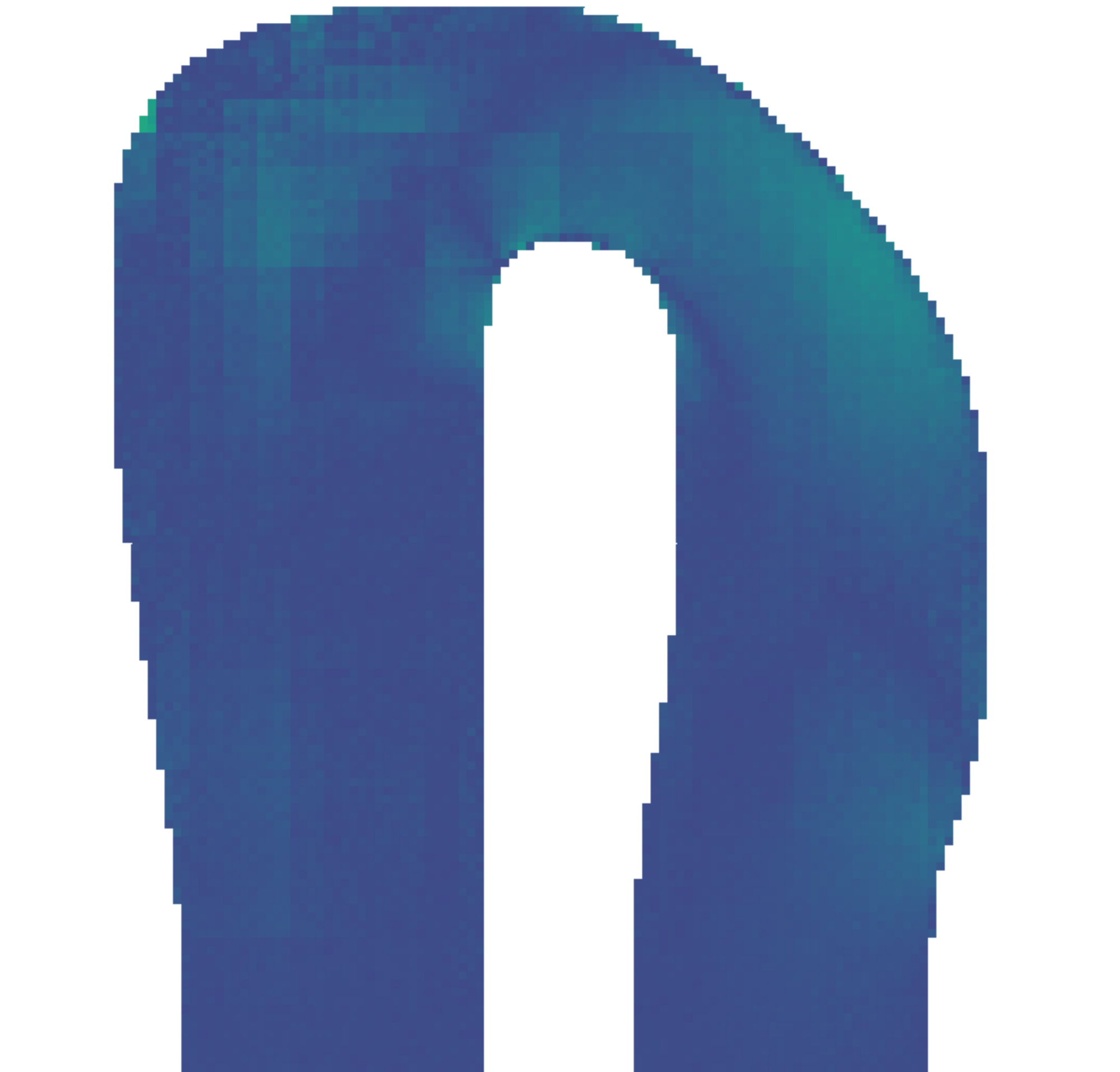}&
        \includegraphics[height=.22\textwidth, keepaspectratio]{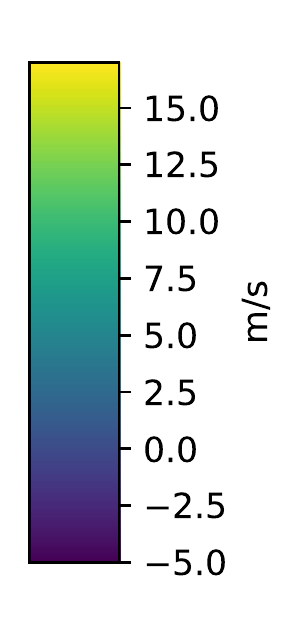}\\
        \includegraphics[width=.2\textwidth, keepaspectratio]{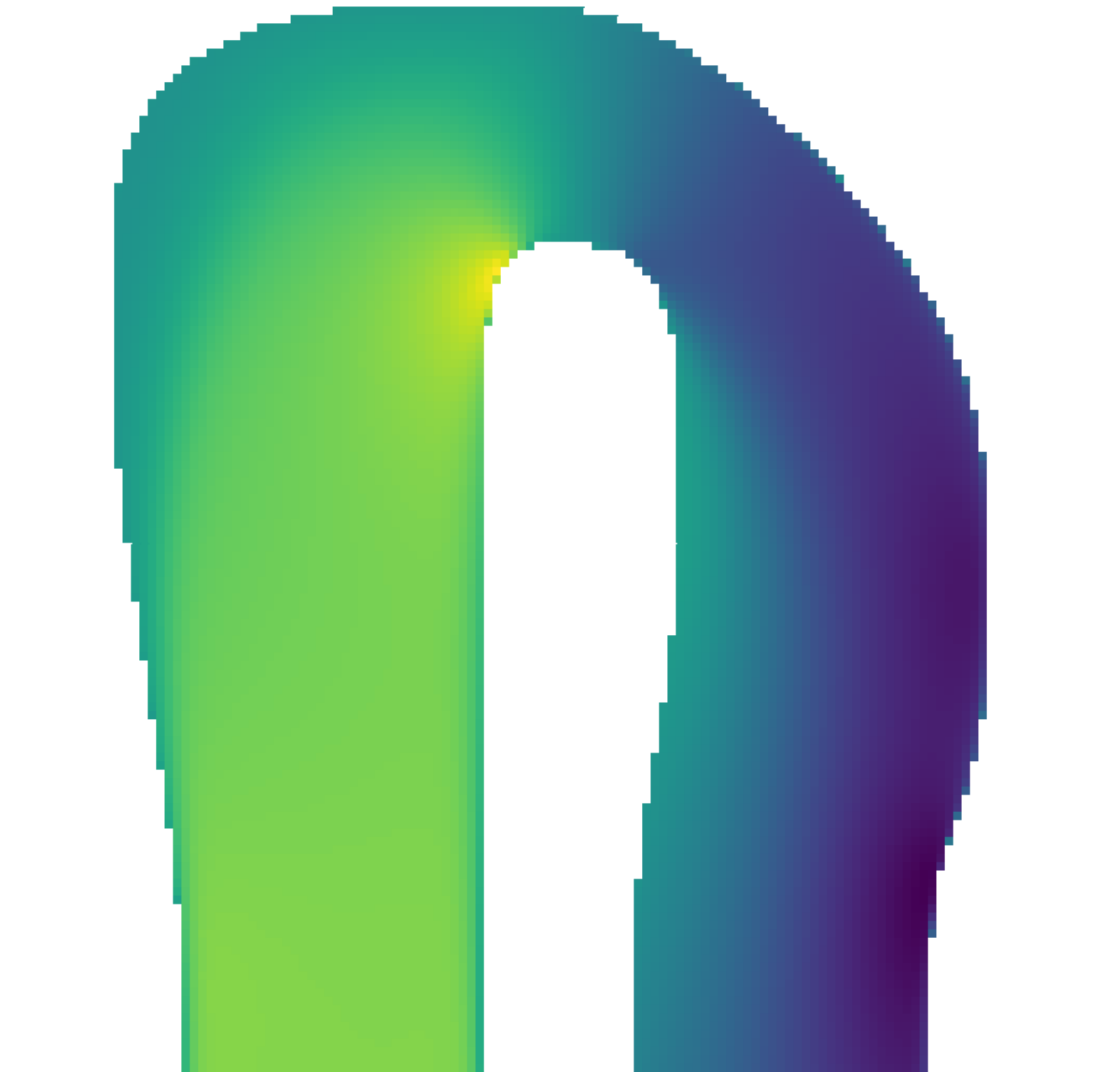}&
        \includegraphics[width=.2\textwidth, keepaspectratio]{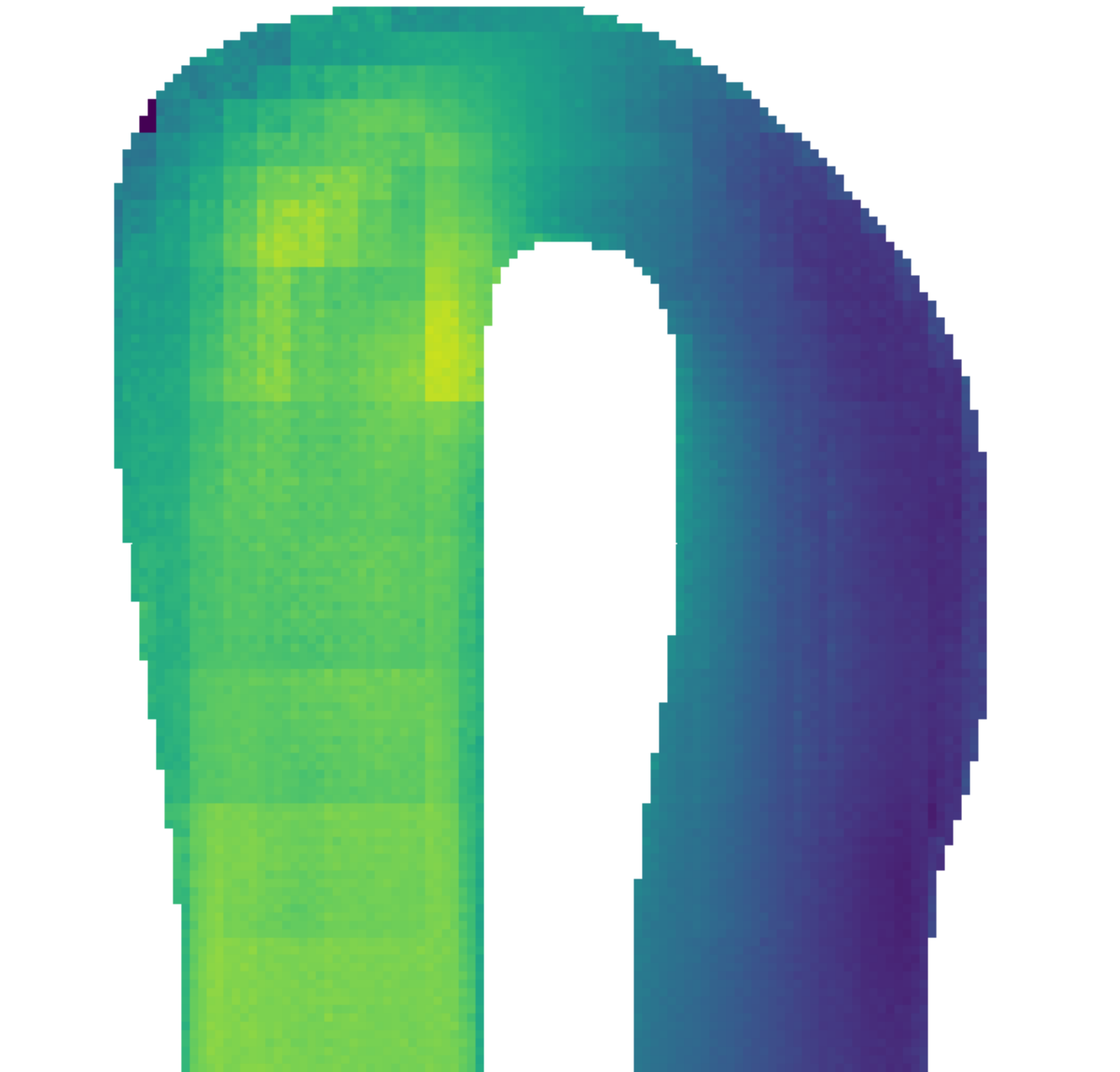}&
        \includegraphics[width=.2\textwidth, keepaspectratio]{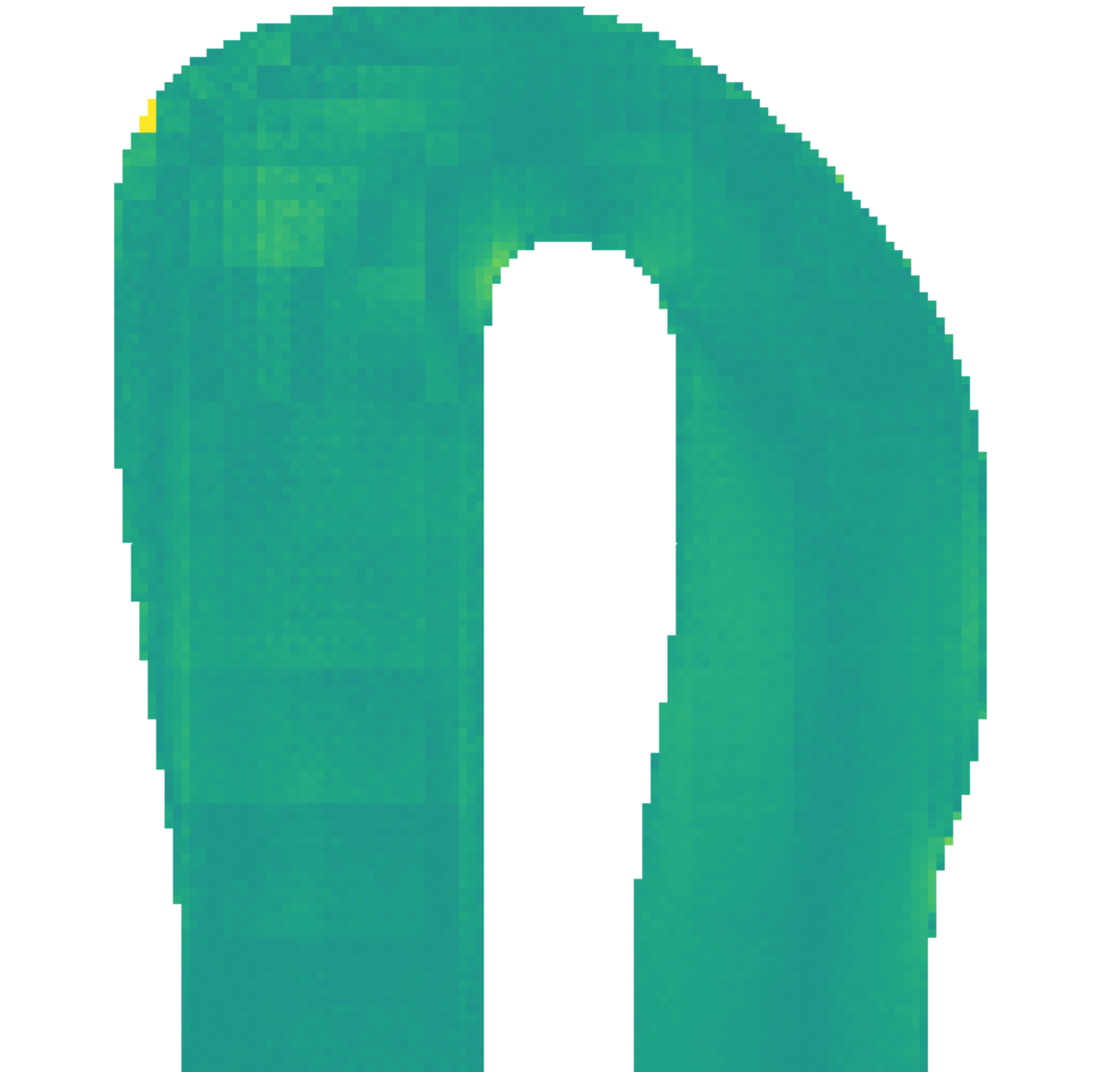}&
        \includegraphics[height=.22\textwidth, keepaspectratio]{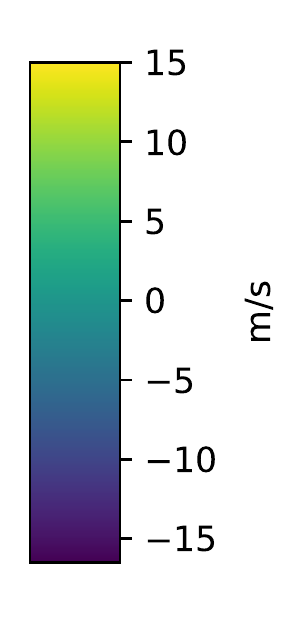}\\
        \includegraphics[width=.2\textwidth, keepaspectratio]{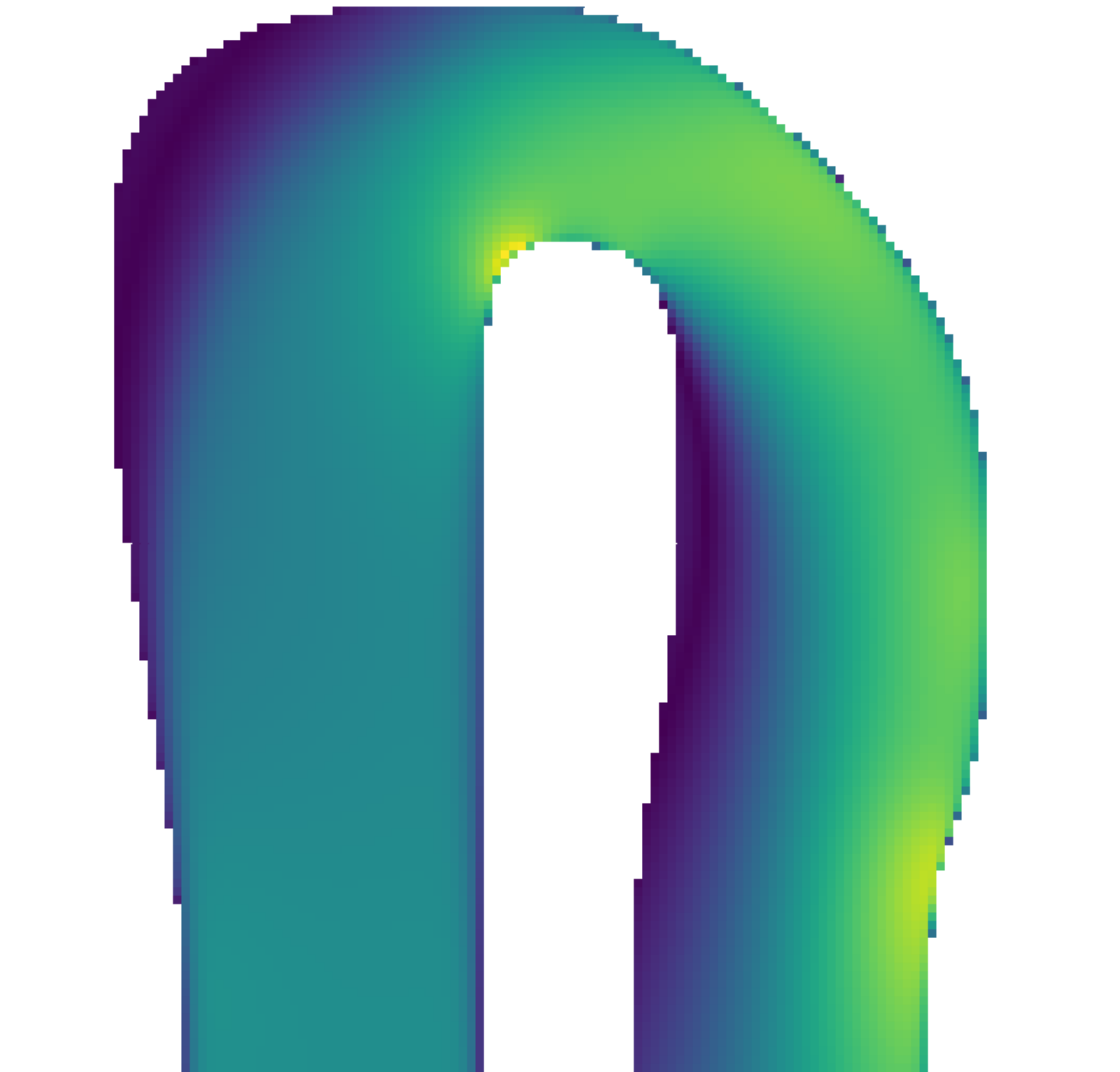}&
        \includegraphics[width=.2\textwidth, keepaspectratio]{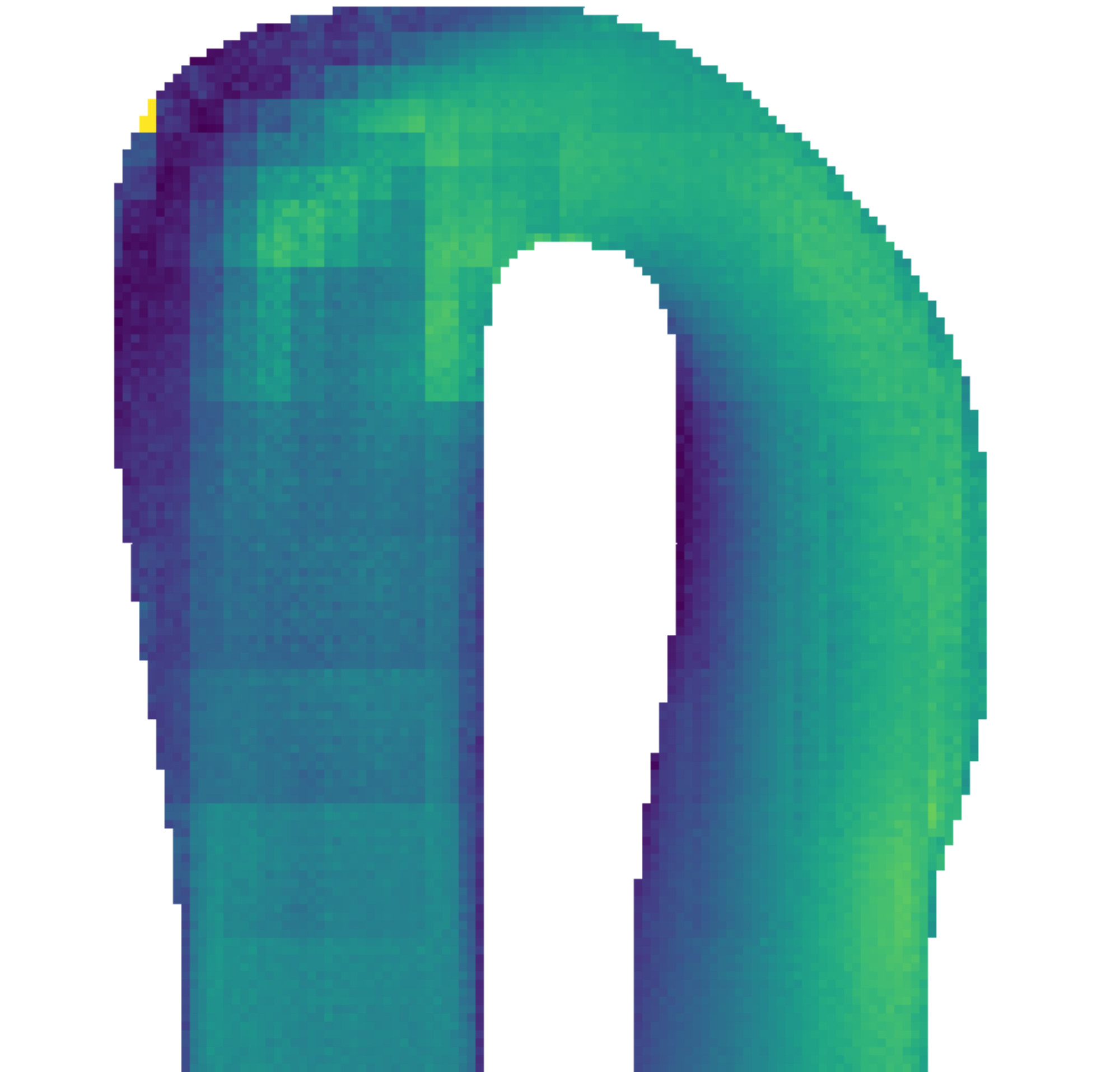}&
        \includegraphics[width=.2\textwidth, keepaspectratio]{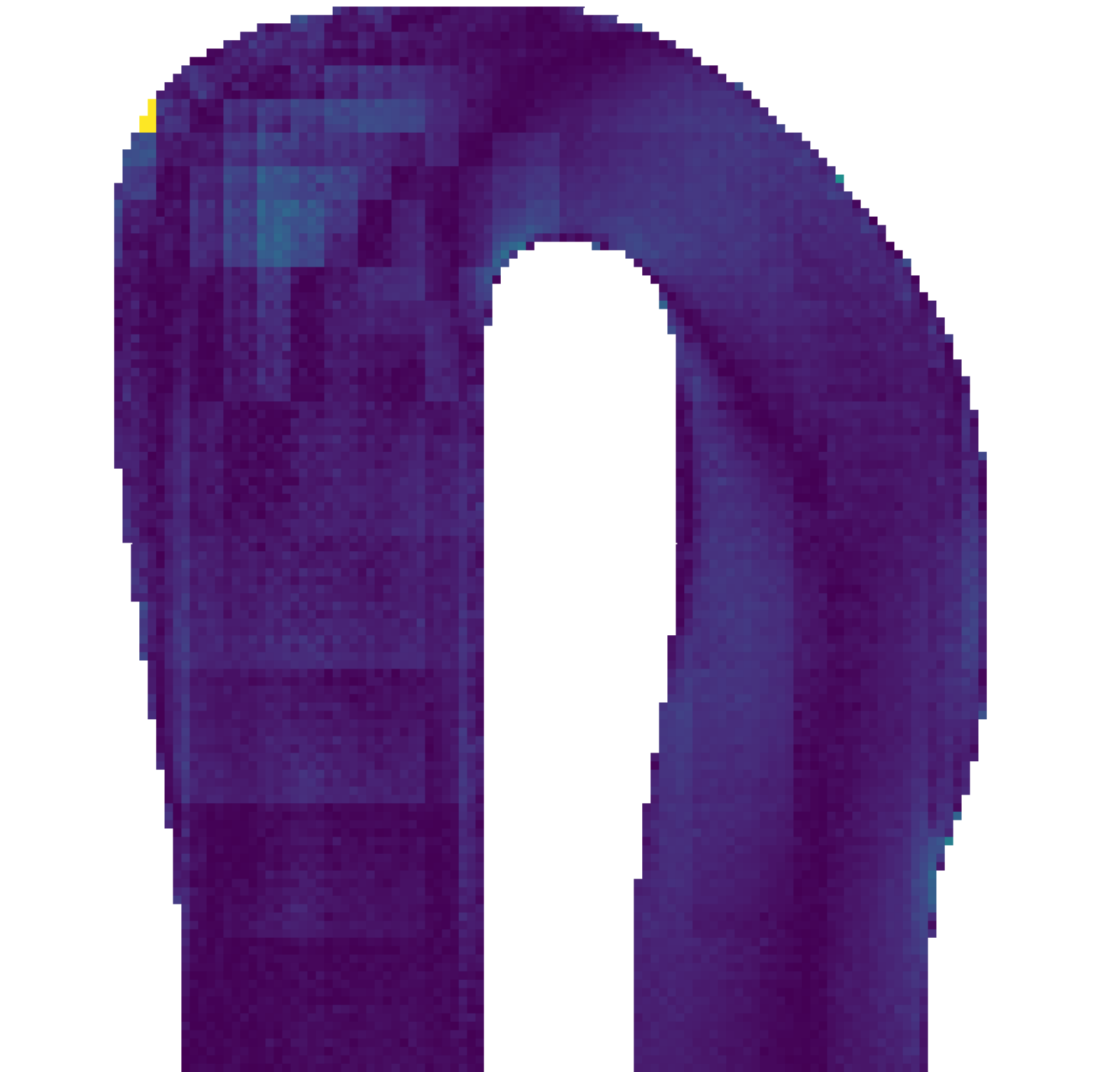}&
        \includegraphics[height=.22\textwidth, keepaspectratio]{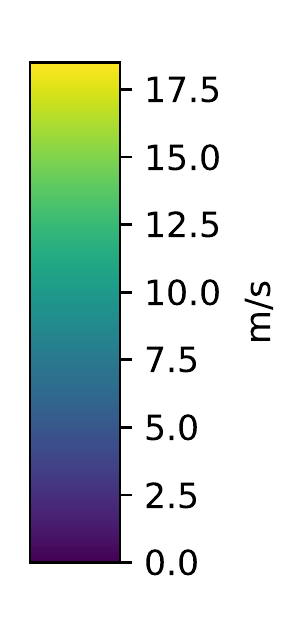}\\
    \end{tabular}
    \caption{Velocity fields belonging to a geometry from the test set. From top to bottom: $v_x$, $v_y$ and $v_{mag}$ fields. From left to right: CFD result, CNN prediction and absolute difference.}
    \label{figres}
\end{figure*}

\nomencNormal[20]{$v$}{velocity}{$m/s$}

\nomencSubSupScript[20]{$x$}{x component of a vector}

\nomencSubSupScript[20]{$y$}{y component of a vector}

\nomencSubSupScript[20]{$mag$}{magnitude of a vector}

\begin{figure}
    \centering
    \includegraphics[width=.45\linewidth, keepaspectratio]{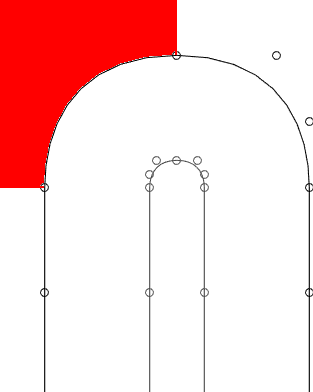}
    \caption{U-bend with the restricted area marked red for the geometries in the training set}
    \label{restricted}
\end{figure}

$4500$, $900$ and $900$ U-bend variants were generated randomly with uniform distribution for training, validation and test sets, respectively. The training and validation data were generated such, that the geometries could not evolve to the upper left corner of the available space. The flow stagnates in this area in most geometry variants if there is a bulge there. Similar flow patterns can develop in other regions of the geometry, so if the CNN generalises well, it should learn to predict the flow field in the restricted area as well. To evaluate the generalization capabilities of the proposed system geometry was allowed to grow in the upper left corner in the case of test sets. The restricted area is depicted in Figure~\ref{restricted}.

Moreover, a data augmentation function was utilized that was allowed to rotate images by 90 degrees counter-clockwise and/or flip them upside down with a probability of $0.44$. The transformation was carried out on both the geometry representation and on the velocity fields. By applying this technique the neural network was forced to gain inference connected to the shape features instead of the absolute location of them. Please note that this type of augmentation makes sense here only because of the restricted type of geometries; it is not a general approach for surrogating CFD models.

Summarising the steps, the experiment was conducted as follows.

\begin{enumerate}
    \item The parameter lists for training, validation and test sets were generated.
    \item The geometries were built and the numerical simulations were performed.
    \item Data were post-processed to get the binary images and signed distance fields as geometry representations.
    \item The hyperparameters of the convolutional neural network were optimised with the training data.
    \item The best neural network topology was trained with the training data.
    \item Tests were performed on the test data.
\end{enumerate}

The signed distance field was chosen as input during the hyperparameter optimisation as it is stated a better performer in \cite{xiao} than binary image. The best topology was trained first with signed distance field input with and without data augmentation, moreover with and without residual connection. Thereafter, the best combination (both data augmentation and residual connection) was trained and tested with binary image input as well. Results are summarised in Table~\ref{results}.

Both data augmentation method and residual connection in the network contributed to a better prediction on the test set. However, it is not clear which geometry representation is better for convolutional neural networks. Treating the geometry as binary image was better in this case but it contradicts the results of \cite{xiao}. Likely, the distance function could not reveal any additional information here compared to binary image as geometries were very similar.

\begin{table}[h]
    \caption{Average prediction RMS error on the test velocity fields. (RC -- residual connection, DA -- data augmentation, BIG -- binary image, SDF -- signed distance field)}
    \label{results}
    \begin{tabularx}{7.4cm}{|X|X|}
        \hline Training method & Prediction error, m/s\\
        \hline SDF, no RC, no DA    & 0.1071\\
        \hline SDF, RC, no DA       & 0.0820\\
        \hline SDF, no RC, DA       & 0.0673\\
        \hline SDF, RC, DA          & 0.0350\\
        \hline BIG, RC, DA          & 0.009\\
        \hline
    \end{tabularx}
\end{table}

The time requirements of the prediction and the simulation are stated in Table~\ref{comptime}. for 1 and for 900\footnote{Such amount of evaluation is common in optimisation-driven design.} evaluations carried out on the hardware described in Section~\ref{techdat}.

\begin{table}[h]
    \caption{Computational time for the CNN and for the CFD.}
    \label{comptime}
    \begin{tabularx}{7.4cm}{|p{5.2cm}|S[table-format=4.3]|}
        \hline Utilized hardware                    & \textrm{time, s}\\
        \hline 1xCNN prediction on CPU              & 0.826\\
        \hline 1xCNN prediction on GPU              & 0.0325\\
        \hline 1xCFD simulation on 4 CPU cores      & 12.6\\
        \hline 900xCNN prediction on GPU            & 7.45\\
        \hline 900xCFD simulation on 4 CPU cores    & 11340.\\
        \hline
    \end{tabularx}
\end{table}

Results of test set are depicted in Figure~\ref{figres}., where such a geometry is shown, where the fluid domain is expanded to the area, that was restricted for the geometries in the training set. Although a small discrepancy is seen in the upper left corner the CNN was able to predict the flow stagnation in the problematic area, implying that the network was able to learn some generalised knowledge about this phenomenon.

\section{Conclusion and outlook}
The presented convolutional neural network was successful in learning abstractions from velocity fields by being able to predict the stagnating flow in a zone that was not seen during the training. The CNN is two orders of magnitude faster than the CFD simulation when just one geometry is evaluated. The gain grows when a larger number of geometries are evaluated as the prediction is highly parallelisable even on a consumer-grade GPU.

This suggests that using CNNs to surrogate CFD models can be advantageous in optimisation tasks, based on some kind of evolution algorithm as a population of geometries can be evaluated at once. In cases, where the accuracy is crucial, the convolutional network could be utilised to warm-start the accurate CFD simulation.

On the other hand, although the presented convolutional neural network performed well on the specific case, its scope of usability is limited owing to the similar patterns presented in the training data. Further work has to aim at improving generalisation to make the benefits of CNN-based surrogate modelling available without excessive preliminary computations.

\section*{Acknowledgements} 
The research presented in this paper has been supported by the European Union, co-financed by the European Social Fund (EFOP-3.6.2-16-2017-00013) and by the BME-Artificial Intelligence FIKP grant of EMMI (BME FIKP-MI/SC). We gratefully acknowledge the support of NVIDIA Corporation with the donation of the Titan Xp GPU used for this research.

\bibliography{cmff}
\end{document}